\documentclass{article}

\usepackage[preprint]{neurips_2023}

\usepackage[utf8]{inputenc} 
\usepackage[T1]{fontenc}    
\usepackage{url}            
\usepackage{booktabs}       
\usepackage{amsfonts}       
\usepackage{nicefrac}       
\usepackage{microtype}      
\usepackage{float}
\usepackage{comment}
\usepackage{graphicx}
\usepackage{amsmath}
\usepackage{bbm}
\usepackage{array}
\usepackage{amssymb}
\usepackage{bbding}
\usepackage{threeparttable}
\usepackage{longtable}
\usepackage[table]{xcolor}
\usepackage{caption}
\usepackage{subfigure}
\usepackage[hang,flushmargin]{footmisc}
\usepackage{multirow}
\usepackage[toc,page]{appendix}
\usepackage{soul}
\usepackage{color}
\usepackage{xcolor}

\usepackage[utf8]{inputenc} 
\usepackage[T1]{fontenc}    
\usepackage{hyperref}       
\usepackage{url}            
\usepackage{booktabs}       
\usepackage{nicefrac}       
\usepackage{microtype}      
\usepackage{lipsum}
\usepackage{fancyhdr}       
\usepackage{graphicx}       
\usepackage{caption}
\graphicspath{{media/}}     

\usepackage{framed}
\usepackage{latexsym}
\usepackage{graphicx}
\usepackage{amsmath,amssymb,amsfonts}
\usepackage{algorithmic}
\usepackage{textcomp}
\usepackage{tabu}
\usepackage{multirow}
\usepackage{multicol}
\usepackage{float}
\usepackage{makecell}
\usepackage{pifont}
\usepackage{xcolor}

\title{UniBrain: Universal Brain MRI Diagnosis with
Hierarchical Knowledge-enhanced Pre-training}

\author{
  Jiayu Lei\textsuperscript{1,2},
  Lisong Dai\textsuperscript{4},
  Haoyun Jiang\textsuperscript{3,2}, 
  Chaoyi Wu\textsuperscript{3,2},
  Xiaoman Zhang\textsuperscript{3,2},
  Yao Zhang\textsuperscript{2} \\
  \textbf{Jiangchao Yao\textsuperscript{3,2,\dag}},
  \textbf{Weidi Xie\textsuperscript{3,2}},
  \textbf{Yanyong Zhang\textsuperscript{1}}, 
  \textbf{Yuehua Li\textsuperscript{4}},
  \textbf{Ya Zhang\textsuperscript{3,2}},
  \textbf{Yanfeng Wang\textsuperscript{3,2,\dag}}  \\ [2pt]
  \tt\small{misslei@mail.ustc.edu.cn}\\ [2pt]
   \textsuperscript{1}University of Science and Technology of China \\[2pt]
   \textsuperscript{2}Shanghai AI Laboratory \\[2pt]
   \textsuperscript{3}Shanghai Jiao Tong University \\[2pt]
   \textsuperscript{4}Shanghai Sixth People’s Hospital Affiliated to Shanghai Jiao Tong University \\[2pt]
   \url{https://github.com/ljy19970415/UniBrain}
}

\begin{document}

\maketitle

\begin{abstract}
Magnetic resonance imaging~(MRI) have played a crucial role in brain disease diagnosis, with which a range of computer-aided artificial intelligence methods have been proposed. 
However, the early explorations usually focus on the limited types of brain diseases in one study and train the model on the data in a small scale, yielding the bottleneck of generalization. Towards a more effective and scalable paradigm, we propose a hierarchical knowledge-enhanced pre-training framework for the universal brain MRI diagnosis, termed as UniBrain. Specifically, UniBrain leverages a large-scale dataset of 24,770 imaging-report pairs from routine diagnostics. Different from previous pre-training techniques for the unitary vision or textual feature, or with the brute-force alignment between vision and language information, we leverage the unique characteristic of report information in different granularity to build a hierarchical alignment mechanism, which strengthens the efficiency in feature learning. Our UniBrain is validated on three real world datasets with severe class imbalance and the public BraTS2019 dataset. It not only consistently outperforms all state-of-the-art diagnostic methods by a large margin and provides a superior grounding performance but also shows comparable performance compared to expert radiologists on certain disease types.

\end{abstract}

\renewcommand{\thefootnote}{}
\footnotetext{\dag: Corresponding author.}

\section{Introduction}
\label{sec:introduction}

Due to the benefit of the non-invasive nature and superior soft tissue contrast \cite{runge2015magnetic,lerch2017studying}, magnetic resonance imaging (MRI) has been considered as a reliable imaging method for brain disease diagnosis. With the rich morphological views of MRI modalities, \textit{e.g.,} T1-Weighted Imaging (T1WI), T2-Weighted Imaging (T2WI), T2-Weighted Fluid Attenuated Inversion Recovery (T2FLAIR), and Diffusion-Weighted Imaging (DWI), clinicians can effectively diagnose a wide range of brain diseases such as glioma, brain hemorrhage, and acute cerebral infarction \cite{katti2011magnetic}. To save human labor and source, the computer-aided way draws more attention with the development of artificial intelligence. 

Recently, deep learning has achieved an impressive performance on brain MRI data. For example, the deep learning models in~\cite{9587756,9103502} have shown non-inferior diagnostic capability to human experts on Alzheimer disease diagnosis or glioma segmentation. Nevertheless, these early works~\cite{jun2021medical,zhou2022self,tang2022self} usually focus on the specific brain diseases of a few types and the models are trained on datasets in a small scale, maintaining a weak generalization on more brain diseases. As the pre-training paradigm becomes prevalent to address this dilemma, how to build a framework for effective and scalable multi-disease diagnosis remains open and draws an increasing attention,~\textit{e.g.,} the recent studies for chest X-ray~\cite{zhang2023knowledge,wu2023medklip}. However, currently, the pre-training models for the multimodal brain MRI data have not yet well considered~\cite{runge2015magnetic,lerch2017studying} .

Towards the pre-training on brain MRI data, there are actually several challenges: 1) \emph{sufficient available data}. The annotation in the field of brain diagnosis is always expensive and time-consuming, which leads to a lack of large-scale datasets that encompass a wide range of brain diseases and MRI modalities; 2) \emph{specific domain knowledge}. The complex nature of brain diseases necessitates a profound understanding of various MRI modalities and their corresponding disease manifestations in a fine-grained granularity, which promotes the efficiency of pre-training; 3) \emph{interpretable prediction}. As the typical diagnosis coexists with visual evidences, it is better for a framework to have some visual grounding along with the prediction, which helps radiologists understand the system and build the trust between humans and machines~\cite{wu2023medklip}.

In this paper, we collect a large-scale brain MRI dataset that contains 24,770 imaging-report pairs of diverse brain diseases, and propose a knowledge-enhanced pre-training framework for the brain MRI diagnosis. Specifically, our design intuition builds upon a specific observation: the brain MRI report is composed of fine-grained components corresponding to each MRI modality, and coarse-grained conclusion of their union. This essentially provides us the potential to construct a hierarchical mechanism to strengthen the pre-training efficiency. Besides, given the diverse disease types and the rich MRI modalities, it is possible to pursue a universal paradigm with the strong generalization by encoding the commonality of brain diseases from both vision and text information. In summary, our contribution can be categorized as follows:

\begin{itemize}
\item[$\bullet$] We propose a novel hierarchical knowledge-enhanced pre-training framework that learns on a large-scale brain MRI dataset to pursue universal brain disease diagnosis.
\item [$\bullet$] We design an automatic report decomposition for the fine-grained vision-language alignment to improve pre-training, and build a coupled perception module that can diagnose any brain disease with the proper description.
\item[$\bullet$] We conduct extensive experiments to verify the effectiveness of the proposed architecture, and compare with many state-of-the-art (SOTA) pre-training algorithms on public BraTS2019 dataset and three in-house datasets.
Our model not only consistently achieves the superior performance to the baselines \textit{w.r.t.} the brain disease diagnosis and grounding but also yields comparable diagnosis performance compared to expert radiologists on certain disease types.
\end{itemize}

The rest of this paper is organized as follows: Section~\ref{Sec2} presents related works. Section~\ref{Sec3} describes in detail the proposed UniBrain pre-training framework. In Section~\ref{experiment section}, extensive experiments on three real world datasets and BraTS2019 are conducted to demonstrate the effectiveness of our pre-training model with ablation study in Section~\ref{ablation study} explains the impact of submodules. Section~\ref{discussion} discusses advantages and limitations of the proposed method. We conclude our work in Section~\ref{conclusion}.

\section{Related Work}
\label{Sec2}

\subsection{Deep Learning in Brain MRI Diagnosis}
In recent years, many deep learning methods have been proposed for the medical applications on brain MRI. According to the brain diseases of interests, existing works can be mainly summarized into three types: 1) \emph{for brain tumors}. The brain tumor segmentation tasks with MRI have been well studied. Many works follow the framework of UNETR~\cite{hatamizadeh2022unetr} and explore the improvement of the encoder structure~\cite{wu2023d,liu2022pc,wyburd2021teds,zhang2022mmformer,li2022transbtsv2,chen2021transunet}. 2) \emph{for stroke}. Automated segmentation and classification of images in the time-critical pathology is critical and practical. In this spirit, deep learning for abnormality detection and segmentation in lacunar and acute cerebral
infarction has drawn much attention~\cite{liu2021deep,zhang2021stroke}. 3) \emph{for Alzheimer’s disease}. Recently, there are several arts devoted to Alzheimer’s disease diagnosis~\cite{jun2021medical,9587756}, which have achieved impressive performance. For example, the deep learning models in~\cite{9587756} have shown non-inferior diagnostic capability to human experts on Alzheimer's disease diagnosis. Despite effectiveness, these studies pay attention to a limited range of brain diseases. In contrast, our UniBrain targets to offer universal diagnosis of diverse brain diseases via a pre-training framework.

\begin{figure*}[t!]
\centerline{\includegraphics[width=\textwidth,height=0.29\textwidth]{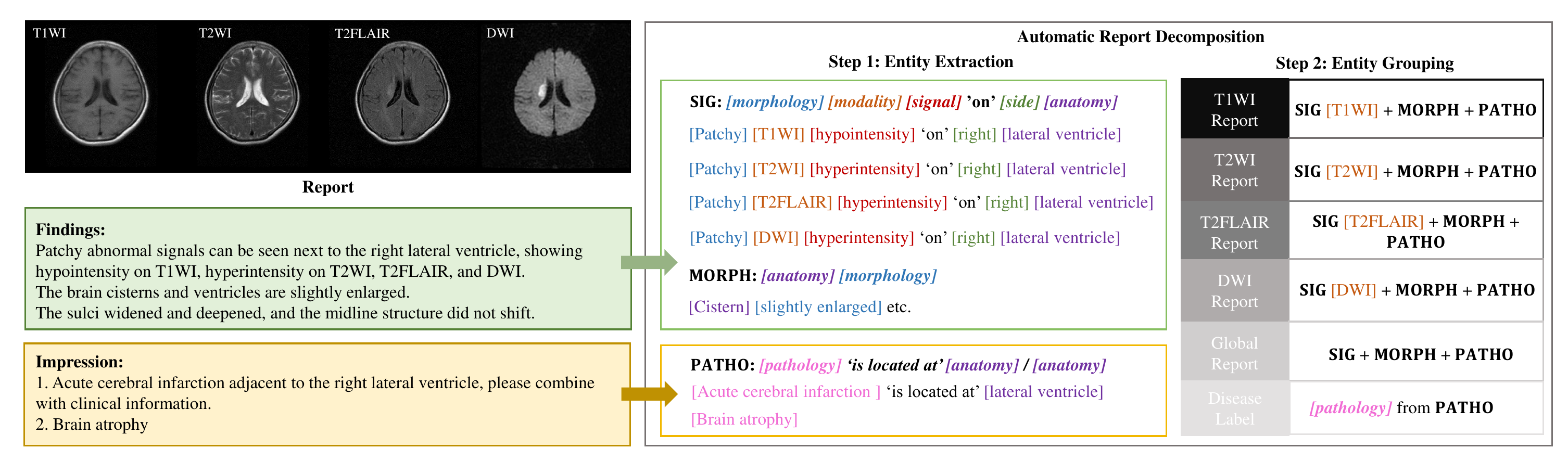}}
\caption{The brain MRI report example and its decomposition. Specifically, we design an automatic report decomposition pipline to extract modality-wise and global MRI report (including the disease labels) from the vanilla MRI report findings and report impression. 
} 
\label{preprocess_report}
\end{figure*}

\subsection{Knowledge-enhanced Pre-training in Medical Domain}
As vision-language pre-training (VLP) has achieved significant performance in the general domain, a range of explorations in the perspective of pre-training have been conducted to improve the performance of medical applications~\cite{wu2023k,xie2021survey}. Generally, these works can be summarized into two categories: the model architecture design and the data augmentation. For the former style, the domain knowledge including the principles of radiology and diagnostics are used to guide the design of the model structure, which makes the learning more efficient on the medical data~\cite{huang2020dual,li2019attention,wang2020learning,fang2019attention,mitsuhara2019embedding}. For the latter style, the auxiliary medical data or description knowledge will be involved into the training~\cite{tan2019expert,xie2018knowledge,hussein2017risk,chen2016automatic,zhang2022contrastive,chauhan2020joint,huang2021gloria,wu2023medklip,zhang2023knowledge}. For instance, MedKLIP~\cite{wu2023medklip} leverages the knowledge disease description to enhance the pre-training and outperform the previous SOTA on zero-shot diagnosis and grounding on chest X-ray. KAD~\cite{zhang2023knowledge} integrated a well-established medical knowledge graph to enhance the self-supervised training, which greatly improves the generalization ability and acquire the SOTA zero-shot results on all public chest X-ray data. Nevertheless, when it comes to the brain diseases, there is a lack of sufficient study of pre-training to show the promise. Especially, the brain MRI data contains multiple modalities, which is different from the chest-xray and requires the specific consideration in design.

\section{Method}
\label{Sec3}
In this section, we will first introduce the preliminary and discuss our motivation in utilizing the report information for pre-training. Then, we will present our automatic report decomposition strategy and hierarchical knowledge-enhanced pre-training framework sequentially.

\subsection{Problem Statement}

Let $D_{\text{train}}=\{(x^1,r^1),...,(x^{N},r^{N})\}$ denote a collection of $N$ image-report pairs, where each $x$ is a collection of $K$ MRI modalities, \textit{i.e.,} $x=(x_1,\dots,x_K)$, and $r$ refers to the corresponding MRI case report. Our goal is to construct a pre-training framework for universal brain disease diagnosis.  

After pre-training, we can infer the likelihood of the brain diseases given a disease query set $Q$, namely, $p=f(x,Q)$.

\subsection{Motivation}\label{report analysis}

Pioneering works have proved the effectiveness of incorporating the radiology report into visual-language pre-training (VLP) in medical field~\cite{zhang2023knowledge,wu2023medklip,huang2021gloria,boecking2022making,zhang2022contrastive}. Nevertheless, previous methods are mainly developed for chest X-ray~\cite{zhang2023knowledge,wu2023medklip,smit2020chexbert,irvin2019chexpert,mcdermott2020chexpert++,peng2018negbio}, whose report corresponds to single chest X-ray imaging. In comparison, brain MRI report contains information for multi-modal imagings, so the VLP methods on unimodal chest X-ray is not efficient for brain MRI. 

We present an example of the brain MRI report in Fig.~\ref{preprocess_report} (left), which consists of two components: \emph{report findings} and \emph{report impression}. As shown in report findings, we can find the information focus on different aspects. For example, signal intensities directly correlate to each modality from the description ``patchy abnormal signals can be seen next to the right lateral ventricle, showing hypointensity on T1WI, hyperintensity on T2WI, T2FLAIR, and DWI" or morphology changes on different anatomies from ``the brain cisterns and ventricles are slightly enlarged" and ``the sulci widened and deepened". In report impression, it provides the disease conclusion and the corresponding located anatomies \textit{e.g.,} ``Acute cerebral infarction adjacent to the right lateral ventricle".

Motivated by this observation in the MRI report, we propose to decompose the report into some structured formats and build a hierarchical knowledge-enhancement for pre-training. 

\begin{figure*}[!t]
\centerline{\includegraphics[width=0.9\textwidth,height=0.43\textwidth]{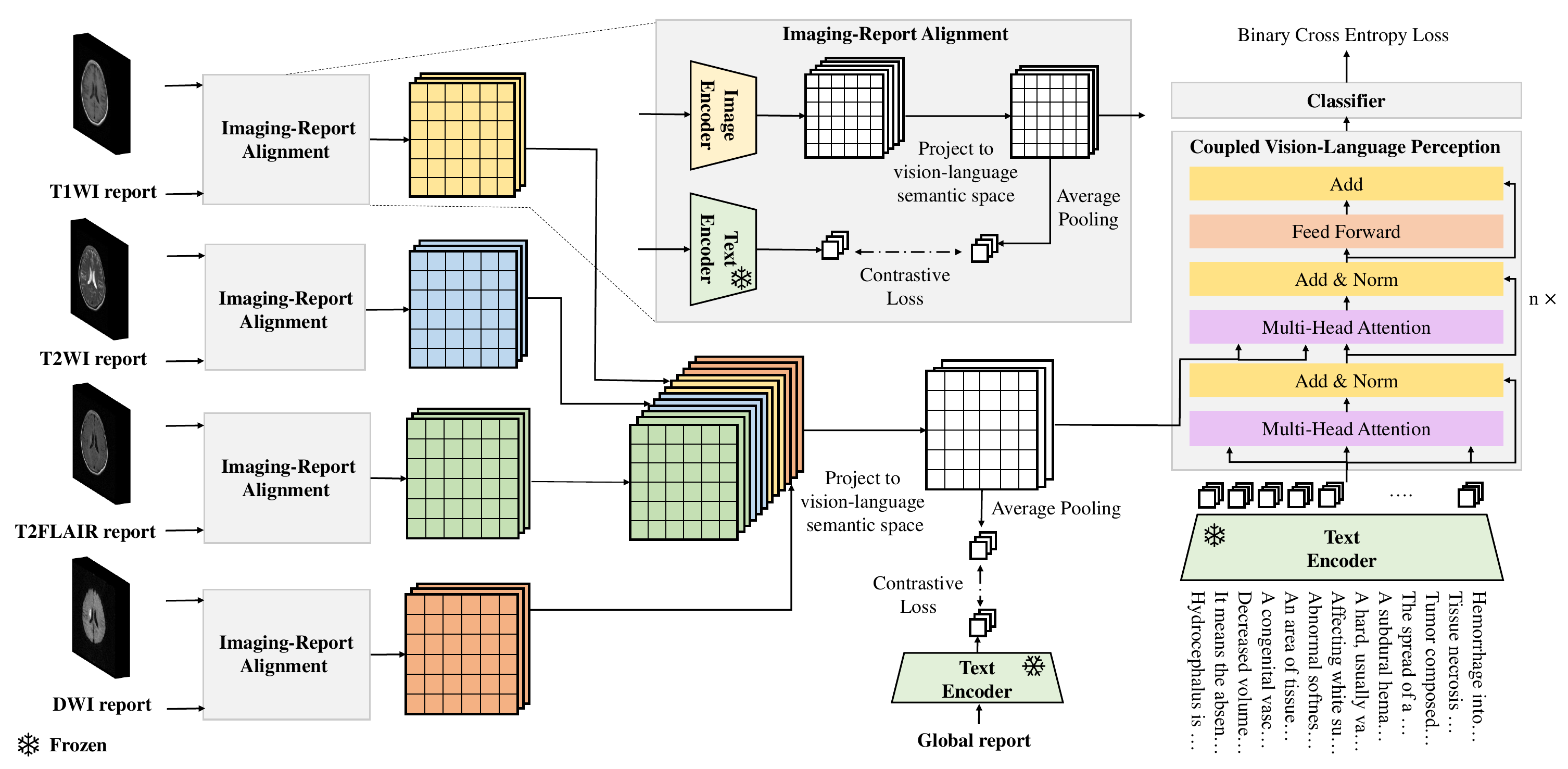}}
\caption{The illustration of our pre-training framework UniBrain. We utilize four transverse MRI modalities, namely T1WI, T2WI, T2FLAIR, DWI, and case reports to train UniBrain. In this framework, we first align the modality-wise imaging-report feature, then we project the concatenated features to vision-language semantic space, and align the global imaging-report feature. Finally, the global image feature acts as the key and value, and the disease description set acts as the query for the coupled vision-language perception module to produce the final multi-class classification.
}
\label{workflow_fig}
\end{figure*}

\subsection{Automatic Report Decomposition} \label{report preprocess}
As shown in Fig.~\ref{preprocess_report}, we propose an automatic report decomposition (ARD) pipline, which consists of two steps: 1) \emph{entity extraction}. Extract the most critical entities according to radiologist experience and reformulate them into structured formats; 2) \emph{entity grouping}. Assign the structured sentences into modality-wise and global reports, and further get the disease labels for classification. Then, the reports will be applied in the proposed pre-training architecture in Fig.~\ref{workflow_fig}.

\subsubsection{Entity Extraction}
In this step, we aim to select informative entities from the report to avoid the interference of extraneous words and complex grammar. Specifically, we follow the radiologist experience to define 6 kinds of entities, \textit{i.e.,} \textit{anatomy}, \textit{side}, \textit{modality}, \textit{signal}, \textit{morphology}, and \textit{pathology}. 
 
Then, given a brain MRI report $r$ with a set of sentences, i.e. $r = \{s_{1},s_{2},...,s_{M}\}$, we independently extract entities for each sentence, and convert the sentence to one of the following structured formats: 1) \texttt{SIG}= \textit{\{[morphology] [modality] [signal] `on' [side] [anatomy]\}}, which indicates the information about modality signal, such as `Patchy DWI hyperintensity on right lateral ventricle'; 2) \texttt{MORPH} = \textit{\{[anatomy] [morphology]\}}, which captures morphological changes on anatomies such as `The sulci widened'; 3) \texttt{PATHO} = \textit{\{[pathology] `is located at' [anatomy]\}}\footnote{If no anatomy is specified, \texttt{PATHO}~follows \textit{\{[pathology]\}}.}, which reflects pathological changes such as `Acute cerebral infarction is located at lateral ventricle'. Note that, when any certain entity type in \texttt{SIG}, \texttt{MORPH}~or \texttt{PATHO}~is absent, we will replace it with an empty string.

\subsubsection{Entity Grouping}
This step correlates each structured sentence with modalities so that we can build a hierarchical (modality-wise and global) enhancement to improve pre-training. Specifically, let $\overline{r} = \{\overline{s}_{1},\overline{s}_{2},...,\overline{s}_{M}\}$ denote the structured report after the \emph{entity extraction} step. For each $\overline{s}_{m}$, if it follows the \texttt{SIG}~format, we then add it to the corresponding modality-wise collection $\overline{r}_k$ following the rule as below:
\begin{equation}
    \overline{r}_k\leftarrow \overline{r}_k\cup\{\overline{s}_m\},\ \text{if~modality~$k$~occurs~in}\ \overline{s}_{m}.
\end{equation}
If it follows \texttt{MORPH}~or \texttt{PATHO}~format, we add it to every modality-wise collection following another rule as below:
\begin{equation}
    \overline{r}_{k}\leftarrow \overline{r}_k\cup\{\overline{s}_m\},\ \forall k\in\{1,2,...,K\}.
\end{equation}
After ARD, we can acquire the structured modality-wise reports $\{\overline{r}_1,\overline{r}_2,\dots,\overline{r}_K\}$ and global report $\overline{r}$, which are used in the following pre-training.

\subsection{Hierarchical Knowledge-enhanced Pre-training} \label{hierarchical knowledge}

Considering the characteristic of brain MRI data, we propose a hierarchical knowledge-enhanced pre-training framework, which consists of three important components: 1) the \emph{modality-wise imaging-report alignment} promotes the modality-wise knowledge efficiently encoded into the modality representation; 2) the subsequent \emph{global imaging-report alignment} strengthens the interaction of multiple modality representations, yielding a more comprehensive merged representation; 3) the final \emph{coupled vision-language perception} module compatible with the universal brain disease, and progressively matches the imaging patches with a disease query to generate the fine-grained grounding and the coarse-grained diagnosis.

The details are as follows. First, regarding the imaging modalities, as the abnormal signal intensities have same definition\footnote{In MRI imaging, hyperintensity always means abnormally higher signal intensities and hypointensity always means abnormally lower signal intensities, which applies to all MRI modalities.}, we use a shared image encoder to extract the modality-wise visual representation. Concretely, we denote the latent visual representation of each modality as follows,

\begin{equation}
\textbf{u}_{k}=\phi_{\text{proj}}(\phi_{\text{image}}(x_{k}))\in\mathbb{R}^{l\times d},\ k\in\{1,2,..,K\},
\end{equation}
where $\phi_{\text{image}}(\cdot)$ follows the structure of ResNet3D~\cite{tran2015learning} and we adopt the output of the penultimate layer as the original image patch embedding, $\phi_{\text{proj}}(\cdot)$ is a Multilayer Perceptron (MLP) to project the original image patch embedding to the vision-language semantic space, $l$ denotes the image patch number and $d$ is the embedding dimension of each image patch.

Regarding the report information, we use a text encoder to extract the modality-wise and global report features as below,
\begin{equation}
\begin{split}
& \textbf{v}_{k}=\phi_{\text{text}}(\overline{r}_{k})\in \mathbb{R}^{d},\ k\in\{1,2,...,K\}, \\
& \textbf{v}=\phi_{\text{text}}(\overline{r}) \in \mathbb{R}^{d}.
\end{split} 
\end{equation}

To leverage the relationship among medical entities, we use a MedKEBERT~\cite{gu2021domain} pretrainined on Unified Medical Language System (UMLS) data~\cite{zhang2023knowledge} as $\phi_{\text{text}}(\cdot)$, and freeze its parameters. 

\subsubsection{Modality-wise Imaging-report Alignment}

In this component, we perform a modality-wise alignment between $\textbf{u}_{k}$ and $\textbf{v}_{k}$ ($k\in\{1,2,..,K\}$) as shown in Fig.~\ref{workflow_fig}. Intuitively, by leveraging modality-wise report reference, the modality-wise vision clues can be sufficiently distilled into the corresponding representation. Given a minibatch of $B$ imaging-report pairs, the modality-wise alignment is conducted in the following 

\begin{equation}
\ell_{k}=\frac{1}{B}\sum_{i=1}^B\frac{1}{\Omega(i)}\left(\ln\frac{e^{\mathbf{\hat{u}}_k^{i\top}\mathbf{v}_k^i/\tau}}{\sum_{j=1}^B e^{\mathbf{\hat{u}}_k^{i\top}\textbf{v}_k^j/\tau}} + \ln\frac{e^{\mathbf{v}_k^{i\top}\mathbf{\hat{u}}_k^i/\tau}}{\sum_{j=1}^B 
 e^{\mathbf{v}_k^{i\top}\mathbf{\hat{u}}_k^j/\tau}}\right),
\label{eq:modalwise}
\end{equation}
where $\Omega(i)$ measures the number of reports same to the $i$-th report in the mini-batch\footnote{After ARD, there are some structured reports that can be totally same. This term guarantees to debias cross positive pairs in the mini-batch if repeated structured reports exists, otherwise $\Omega(i)=1$.}, $\mathbf{\hat{u}}_k^i=\phi_{\text{pool}}(\mathbf{u}_k^i)\in\mathbb{R}^d$ that pools the embedding along the patch dimension for each $k$-th modality imaging, and $\tau$ is the learnable temperature parameter. Note that, we follow the CLIP loss~\cite{radford2021learning} to build a formulation of bidrectional constrast learning, namely, one image-to-text contrast term and one text-to-image contrast term in Eq.~\eqref{eq:modalwise}.

\subsubsection{Global Imaging-report Alignment}
In this global imaging-report alignment, we first summarize the  modality-wise feature evidences, \textit{i.e.,} $\{\textbf{u}_{1},\textbf{u}_{2},...,\textbf{u}_{K}\}$, to acquire a global feature vector $\textbf{u}$, which is modeled by the the following projection
\begin{equation}
\begin{aligned}
\textbf{u} = \phi_{\text{fuse}}(\textbf{u}_{1},\textbf{u}_{2},...,\textbf{u}_{K})\in\mathbb{R}^{l\times d}.
\end{aligned}
\end{equation}
where $\phi_{\text{fuse}}(\cdot)$ is a linear neural layer mapping one input in $\mathbb{R}^{l\times K\times d}$ to one output in $\mathbb{R}^{l\times d}$. Then, together with the previous global report embedding $\textbf{v}$, we can perform a contrastive learning similar to Eq.~\eqref{eq:modalwise} but in a global view. The loss for the global imaging-report alignment can be formulated as follows

\begin{equation}
\begin{aligned}
\ell_\text{g}=\frac{1}{B}\sum_{i=1}^B\frac{1}{\Omega(i)}\left(\ln\frac{e^{\mathbf{\hat{u}}^{i\top}\mathbf{v}^i/\tau}}{\sum_{j=1}^B e^{\mathbf{\hat{u}}^{i\top}\textbf{v}^j/\tau}} + \ln\frac{e^{\mathbf{v}^{i\top}\mathbf{\hat{u}}^i/\tau}}{\sum_{j=1}^B 
 e^{\mathbf{v}^{i\top}\mathbf{\hat{u}}^j/\tau}}\right),
\end{aligned} \label{eq:global}
\end{equation}
where $\Omega(i)$ is same to that in Eq.~\eqref{eq:modalwise} and $\mathbf{\hat{u}}^i=\phi_{\text{pool}}(\mathbf{u}^i)\in\mathbb{R}^d$ that pools the global embedding along the patch dimension. By performing such a global alignment, we can efficiently learn complex disease patterns beyond the modal-wise evidences.

\subsubsection{Coupled Vision-Language Perception} \label{CAM}
Regarding the diagnosis component, instead of the straightforward discrminative prediction, we adopt a coupled vision-language perception module (CVP) via a transformer decoder-based structure~\cite{vaswani2017attention} as shown in Fig.~\ref{workflow_fig}. Specifically, we use the global imaging embedding $\textbf{u}$ as key, value and use the pre-trained text encoder mentioned in previous sections to encode the disease descriptions $\textbf{Q}$ as queries. The disease-attentive embedding for the brain MRI on the disease set $\textbf{Q}$ is generated as follows 
\begin{equation}
\mathbf{h} = \phi_{\text{CVP}}(\textbf{u},\phi_{\text{text}}(\textbf{Q})) \in \mathbb{R}^{C\times d},
\end{equation}
where $Q=\{q_{1},..,q_{C}\}$ is a collection of disease descriptions generated by UMLS, $C$ is the number of queried diseases, and $ \phi_{\text{CVP}}(\cdot, \cdot)$ is the transformer decoder. Note that, such a design enjoys two merits: 1) we can apply any type of brain disease if it can be expressed by UMLS with the proper descriptions. This aligns our goal about the universal brain diagnosis. 2) In the progressive transformation with the attention mechanism, the response of each image patch can be explicitly characterized, which provides us some evidence about some specific disease. Specially, we can visualize the class activation map (CAM)~\cite{zhou2016learning} on the latent patch features to make the diagnosis attribution. Finally, $\mathbf{h}$ is fed to a classifier to output a prediction $p\in\mathbb{R}^C$ for $C$ diseases. The classification loss is formulated in the following
\begin{equation}\label{eq:classification}
\ell_{\text{bce}} = \frac{1}{B}\sum_{i=1}^B \frac{1}{C}\sum_{c=1}^C \left(y_c^i\ln p_c^i + (1-y_c^i)\ln (1-p_c^i)\right),
\end{equation}
where $\left[p_1, p_2, \dots, p_C\right]=\phi_{\text{classifier}}(\mathbf{h})$ modeled by an MLP layer, and $y^i$ reflects whether the report impression has the corresponding pathology.

\subsubsection{Training, Inference and Beyond}
In the aforementioned subsections, we proposed three losses corresponding to different roles in hierarchical knowledge enhanced pre-training. Putting all together, our training objective is formulated as
\begin{equation} \label{eq:overall}
\mathcal{L} = \ell_{\text{bce}} + \frac{1}{K+1}\left(\ell_\text{g} + \sum_{k=1}^K \ell_k\right). 
\end{equation}
Note that, in the above equation, we have not introduced hyperparameters to balance different loss terms. This is because that we find maintaining them equally can work well. We tried to balance loss terms with different weights, which is comparable or not better than Eq.~\eqref{eq:overall}. During inference phase, we can directly manipulate the diseases of interest with the proper description in $Q$, and forward to UniBrain to have the predictions. Besides, as our $\phi_\text{CVP}$ is based on transformer decoder structure, one can input queries of any length without changing the pre-trained weights of UniBrain, making it easy for zero-shot diagnosis on unseen categories. When fine-tuning on the downstream dataset, one can accordingly modify the disease query set and train the networks with or without $\ell_\text{g}$ and $\ell_k$ depending on whether the report is available. The results in the following section will sufficiently demonstrate UniBrain can generalize well under open-class and domain shift.

\section{Experiments} \label{experiment section}
To evaluate our method, we conduct experiments on four diverse
datasets of which three are real world medical imaging datasets, one is public dataset BraTS2019. We compare the method against multiple baselines to verify the diagnosis and generalization performance of UniBrain.

\subsection{Datasets and Evaluation Metrics}

\subsubsection{Pre-training Dataset}
\begin{itemize}
\item[$\bullet$] \textbf{Shanghai Sixth People's Hospital Dataset (SSPH)}: We collected 24,770 brain MRI imaging-report pairs at Shanghai Sixth People's Hospital in the study\footnote{This study has been approved by the Ethics Committee of Shanghai sixth people's Hospital [IRB code: 2023-KY-082 (K)].}. Here, all MRI scans were performed on one of the eight MRI scanners (GE Medical Systems Signa Pioneer, Philips Medical Systems EWS, Philips Medical Systems Ingenia, Siemens Prisma, Siemens Skyra, Siemens Verio, UIH uMR 780, UIH uMR 790) from January 2019 to March 2023 at Shanghai Sixth People's Hospital. Each data has a case report and four transverse MRI modalities: T1WI, T2WI, T2FLAIR and DWI. The dataset has 13 categories: normal, lacunar cerebral infarction, brain atrophy, white matter lesions, acute cerebral infarction, chronic cerebral infarction, metastasis, brain hemorrhage, epidura and subdural hemorrhage, meningioma, hemangioma, glioma and hydrocephalus. The category distribution is severely imbalanced as shown in Fig.~\ref{longtail}. Note that, such an imbalanced distribution is common in real-world hospital data due to the disease morbidity and the specialization of hospitals in diagnosing diseases.
\end{itemize}

\subsubsection{Downstream Datasets}
\begin{itemize}
\item[$\bullet$] \textbf{BraTS2019 Dataset}: We use the BraTS2019 dataset~\cite{DBLP:conf/brainles-ws/2019-2} that has four MRI modalities: T1WI, T2WI, T2FLAIR, and T1 contrast-enhanced (T1CE)\footnote{This modality has not  occurred on the SSPH dataset.}, to test the zero-shot or finetuning performance of UniBrain. There are 259 volumes of high-grade glioma (HGG) and 73 volumes of low-grade glioma (LGG), which can be directly used for zero-shot evaluation. Regarding the finetuning setting, we split BraTS2019 into training and testing under the ratio of 7:3 following the way in~\cite{chatterjee2022classification}.
\item[$\bullet$] \textbf{Affiliated Hospital of Nantong University Dataset (AHNU)}: This dataset is collected from Affiliated Hospital of Nantong University, consisting of 706 MRI imagings under 4 modalities: T1WI, T2WI, T2FLAIR, DWI. It contains 13 categories same to those in our pre-training dataset. We mainly use this dataset to test the generalization ability of UniBrain under domain shift.
\item[$\bullet$] \textbf{Wuhan Hankou Hospital Dataset (WHH)}: It is collected from Wuhan Hankou Hospital, which has 618 MRI imagings of 13 categories in SSPH under 4 modalities: T1WI, T2WI, T2FLAIR, DWI. We use this dataset to test the generalization ability of UniBrain under domain shift.
\end{itemize}

\subsubsection{Evaluation Metrics}

We respectively use area under curve (AUC), Accuracy (ACC), Average Precision (AP) and F1-score (F1) for each category and their average number, \textit{i.e.} average AUC (aAUC), average Accuracy (aACC), average F1-score (aF1) and mean Average Precision (mAP), as our evaluation metrics. These metrics are calculated as follows:
\begin{equation}
\begin{aligned}
\text{aAUC} &=\frac{1}{C}\sum_{c=1}^{C}\text{AUC}_{c}, \\
\text{aACC} &=\frac{1}{C}\sum_{c=1}^{C}\frac{\text{TP}_{c}+\text{TN}_{c}}{\text{TP}_{c}+\text{FP}_{c}+\text{TN}_{c}+\text{FN}_{c}},\\
\text{aF1}  &=\frac{1}{C}\sum_{c=1}^{C}\frac{2\text{TP}_{c}}{2\text{TP}_{c}+\text{FP}_{c}+\text{FN}_{c}}, \\
\text{mAP}  &=\frac{1}{C}\sum_{c=1}^{C}\text{AP}_{c}, \\ 
\end{aligned}
\end{equation}
where $\text{TP}_{c}$, $\text{FN}_{c}$, $\text{TN}_{c}$, $\text{FP}_{c}$ denote number of true positive, false negative, true negative, false positive samples for the $c$-th category respectively, and $\text{AUC}_{c}$, $\text{AP}_{c}$ are the common AUC performance and AP performance of the $c$-th category that can be easily calculated in the standard python API.

\subsection{Pre-training Details}

\begin{figure}
\begin{minipage}[htbp]{.5\linewidth}
\centering
\includegraphics[width=\columnwidth]{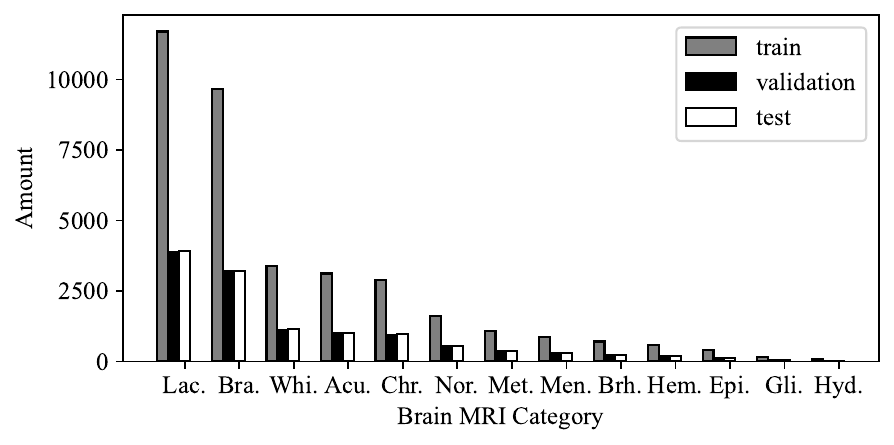}
\caption{The category distribution of the train subset, the validation subset and the test subset in SSPH. The first three letters are used as the shorthand of each class except for “Brain Atrophy” and “Brain hemorrhage” that we denote by "Bra" and "Brh" respectively.}
\label{longtail}
\end{minipage}
\begin{minipage}[htbp]{.5\linewidth}
\centering
\setlength{\tabcolsep}{.015\linewidth}{
\begin{tabular}{lcccc}
\toprule
Method & 2D/3D & \makecell[c]{Data \\ Type} & \makecell[c]{Multi-\\Modality} & VLP \\
\hline
ConVIRT~\cite{zhang2022contrastive} & 2D & X ray & & \ding{52} \\
CheXZero~\cite{tiu2022expert} & 2D & X ray &  & \ding{52} \\
MedKLIP~\cite{wu2023medklip} & 2D & X ray &  & \ding{52} \\
KAD~\cite{zhang2023knowledge} & 2D & X ray &  & \ding{52} \\
TransMed~\cite{dai2021transmed} & 3D & MRI & \ding{52} & \\
DSM~\cite{chatterjee2022classification} & 3D & MRI &  & \\
\hline
UniBrain & 3D & MRI & \ding{52} & \ding{52} \\
\bottomrule
\end{tabular}}
\captionof{table}{Setting Summary of UniBrain and baselines. The setting includes 2D/3D inputs, Data Type, Multi-modality, VLP. UniBrain is the first work using visual-language pre-training on 3D multi-modality brain MRI data.}
\end{minipage}
\end{figure}

SSPH is split into 3:1:1 for training, validation, and test, and each modality input is resized to 224 × 224 × 24. For all our experiment, 3DSeg-8 pre-trained ResNet3D-34~\cite{chen2019med3d} is used as the default image encoder of UniBrain. We will also discuss the impact of the image encoder in Section~\ref{ablation study}. 
The learnable temperature parameter $\tau$ is intialized the same as~\cite{zhang2023knowledge}. 
We set the number of Transformer decoder blocks in $\phi_{\text{CVP}}$ as 4 with 4 heads. For baselines in the experiments, we strictly followed the architectures described in the original papers.

\begin{table*}
\centering
\caption{
Comparison of UniBrain with the baselines in terms of aAUC, aACC, aF1 and mAP. We use the first three letters to represent each class except for “Brain Atrophy” and “Brain hemorrhage” that we use "Bra" and "Brh" respectively. Numbers within parentheses indicate 95\% confidence intervals (CI) and the best results for each column are marked in bold.}
\label{maintable}
\setlength{\tabcolsep}{3pt}
\resizebox{\linewidth}{!}{
\begin{tabular}{c|p{0.8cm}<{\centering}p{0.8cm}<{\centering}p{0.8cm}<{\centering}p{0.8cm}<{\centering}p{0.8cm}<{\centering}p{0.8cm}<{\centering}p{0.8cm}<{\centering}p{0.8cm}<{\centering}p{0.8cm}<{\centering}p{0.8cm}<{\centering}p{0.8cm}<{\centering}p{0.8cm}<{\centering}p{0.8cm}<{\centering}|p{1.1cm}<{\centering}p{1.1cm}<{\centering}p{1.1cm}<{\centering}p{1.1cm}<{\centering}}
\hline
Method & Lac. & Bra. & Whi. & Acu. & Chr. & Nor. & Met. & Brh. & Epi. & Men. & Hem. & Gli. & Hyd. & aAUC & aACC & aF1 & mAP \\
\hline
ConvVIRT~\cite{zhang2022contrastive} & 46.23 & 72.75 & 74.10 & 65.20 & 32.79 & 93.53 & 62.39 & 70.23 & 48.13 & 61.28 & 63.90 & 83.93 & 51.08 & 63.50 & 79.54 & 38.57 & 28.87\\
CheXZero~\cite{tiu2022expert} & 34.45 & 86.71 & 78.53 & 41.59 & 30.83 & 87.95 & 67.94 & 62.66 & 55.06 & 60.85 & 57.67 & 79.81 & 85.66 & 63.83 & 76.63 & 36.16 & 27.51\\
TransMed~\cite{dai2021transmed} & 83.81 & 91.25 & 85.91 & 70.97 & 77.41 & 91.10 & 56.63 & 57.19 & 62.09 & 59.48 & 54.83 & 69.00 & 67.77 &  71.34 & 81.48 & 36.08 & 32.77\\
MedKLIP~\cite{wu2023medklip} & 88.06 & 91.22 & 91.33 & 90.56 & 81.86 & 95.24 & 77.82 & 82.89 & 78.33 & 75.61 & 68.66 & 81.26 & 88.18 & 83.93 & 89.17 & 48.87 & 45.88\\
KAD~\cite{zhang2023knowledge} & 87.88 & 92.46 & 92.67 & 90.25 & 81.85 & 95.68 & 80.76 & 85.37 & 80.60 & 74.27 & 74.27 & 68.94 & 89.56 & 85.57 & 90.61 & 52.09 & 50.33\\
\hline
\textbf{UniBrain} & \textbf{90.35} &  \textbf{94.48} &  \textbf{94.27} &  \textbf{96.1} &  \textbf{88.17} &  \textbf{96.61} &  \textbf{89.32} &  \textbf{91.43} &  \textbf{93.50} & \textbf{79.13} &  \textbf{73.1} &  \textbf{93.75} &  \textbf{97.63} & \textbf{90.71} & \textbf{92.62} & \textbf{62.27} & \textbf{63.27}\\
\multirow{2}{*}{(95\% CI)} & (89.39, 91.30) & (93.85, 95.10) & (93.63, 94.91) & (95.42, 96.78) & (86.93, 89.40) & (96.09, 97.13) & (87.57, 91.07) & (89.27, 93.58) & (91.09, 95.90) & (76.46, 81.80) & (69.53, 76.67) & (89.74, 97.76) & (96.12, 99.13) & (90.11, 91.31) & (92.08, 93.16) & (60.71, 63.82) & (61.42, 65.11) \\
\hline
\end{tabular}
}
\end{table*}

Our UniBrain is implemented in Pytorch and trained with 2 NVIDIA A100 GPUs (each has 80GB memory) for 100 epochs from scratch with a batch size of 16. We adopt the Adam optimizer and its initial learning rate is set to 0.0002 with a poly learning rate schedule, in which the initial rate decays by each epoch with a power 0.9. For data augmentation, we use the following operations: (1) random mirror fipping across the axial, coronal and sagittal planes by a probability of 0.5; (2) random intensity shift in [-0.1, 0.1] and scale in [0.9, 1.1]. The L2 Normalization is also applied for model regularization with a weight decay rate of 1e-5.

\subsection{Baselines}
We consider a range of baselines in comparison: 1) multi-modal MRI classification model: TransMed~\cite{dai2021transmed}; 2) VLP models: ConVIRT~\cite{zhang2022contrastive}, CheXZero~\cite{tiu2022expert}, MedKLIP~\cite{wu2023medklip}, KAD~\cite{zhang2023knowledge}, which are the most advanced methods that apply VLP to medical domain. For the comparison on BraTS2019, we also include DSM~\cite{chatterjee2022classification} which achieves the best result on tumor classification on BraTS2019. Each method is introduced in the following and summarized in Table~\ref{muc}.

\begin{itemize}
\item[$\bullet$] \textbf{ConVIRT}~\cite{zhang2022contrastive} learns the medical visual representations through bidirectional contrast from the image-text pairs.
\item[$\bullet$] \textbf{CheXZero}~\cite{tiu2022expert} fine-tunes the pre-trained CLIP model with the image-text pairs in the medical domain.
\item[$\bullet$] \textbf{MedKLIP}~\cite{wu2023medklip} incorporates the specific medical domain knowledge to enhance the pre-training process.
\item[$\bullet$] \textbf{KAD}~\cite{zhang2023knowledge} utilizes the medical knowledge graph to promote the VLP on chest X-ray data.
\item[$\bullet$] \textbf{TransMed}~\cite{dai2021transmed} combines CNN and Transformer to improve the parotid gland tumor classification through multi-modal MRI imagings.
\item[$\bullet$] \textbf{DSM}~\cite{chatterjee2022classification} uses two spatiotemporal models to classify different types of brain tumours, and has shown a promising and competitive performance on the BraTS2019 dataset.
\end{itemize}

\subsection{Quantitative Results}

Here, we first use the SSPH training set as the pre-training data for UniBrain and other baseline methods~\cite{zhang2022contrastive,tiu2022expert,dai2021transmed,wu2023medklip,zhang2023knowledge}, and test their aAUC, aACC, aF1, and mAP on the SSPH test set. The results presented in Table~\ref{maintable} demonstrate the superior performance of UniBrain across all metrics examined.
\subsubsection{UniBrain outperforms the SOTA diagnosis model}
When compared to TransMed, another multi-modal MRI diagnosis model, UniBrain achieves significant improvements in aAUC, aACC, aF1, and mAP by 19.37\%, 11.14\%, 26.19\%, and 30.5\%, respectively. When compared to the SOTA VLP model KAD, our method outperforms it by 5.14\%, 2.01\%, 10.18\%, and 12.94\% in aAUC, aACC, aF1, and mAP, respectively. These results highlight the advantages of using multi-modal data in a hierarchical enhancement manner for pre-training.
\subsubsection{UniBrain is robust to the imbalanced dataset} 
What is particularly noteworthy regarding UniBrain is its substantial improvement on aF1 and mAP when compared to the SOTA models. This shows the robustness of UniBrain when the dataset is quite imbalance like Fig.~\ref{longtail}. In particular, UniBrain achieves significant improvements in eight minority disease types including acute cerebral infarction, chronic cerebral infarction, metastasis, brain hemorrhage, epidural and subdural hemorrhage, meningioma, glioma, and hydrocephalus, which comprise less than 20\% of the entire dataset.

\begin{figure*}[!t]
\centerline{\includegraphics[width=\linewidth,height=0.164\columnwidth]{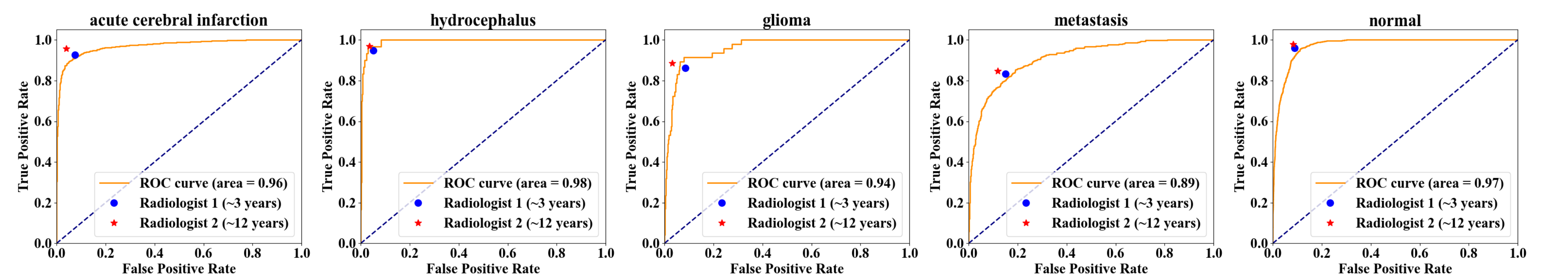}}
\caption{UniBrain achieves the comparable performance on five categories, namely, acute cerebral infarction, glioma, hydrocephalus, metastasis and normal, compared to two brain MRI radiologists with 3 and 12 years of clinical experience respectively.
}
\label{real}
\end{figure*}
\begin{figure*}[!t]
\centerline{\includegraphics[width=\linewidth,height=0.4\linewidth]{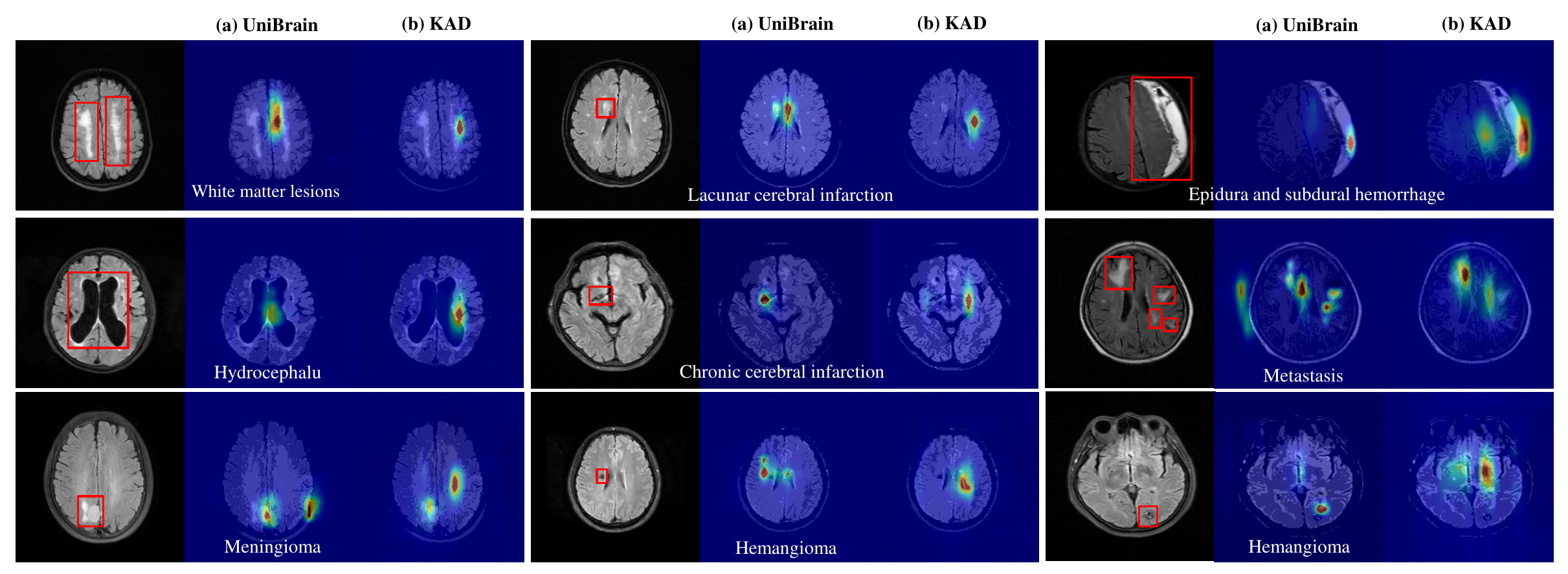}}
\caption{We present the original image (left) and the corresponding attention map generated by UniBrain (column (a)) and KAD (column (b)) respectively. In the original images, lesion areas annotated by radiologists are highlighted by red boxes. In the attention maps, a color spectrum ranging from red to blue is overlaid on the original image, where red represents regions of high attention and blue represents regions of low attention.
}
\label{local}
\end{figure*}
\subsubsection{UniBrain is comparable to radiologists on certain categories} We invite two brain MRI radiologists with 3-year and 12-year clinical experience respectively as our human baseline on 13 categories. We find that UniBrain achieves the comparable results on 5 categories, namely, acute cerebral infarction, glioma, hydrocephalus, metastasis and normal compared to the radiologists as shown in Fig.~\ref{real}. This result demonstrates the great potential for UniBrain to apply to the clinical diagnosis.

\subsection{Qualitative Results}

In Fig.~\ref{local}, we visualize the classification clues by illustrating the original image and the corresponding attention map generated by UniBrain (shown in column (a)) and KAD (shown in column (b)) respectively. As can be seen, in the original images, lesion areas annotated by radiologists are highlighted by red boxes. In the attention maps, a color spectrum ranging from red to blue is overlaid on the original image, where red represents regions of high attention and blue represents regions of low attention. Notably, we observed that when different disease queries were inputted, compared to KAD, UniBrain generated attention maps that focused on disease-specific lesions more accurately. This grounding performance suggests that our hierarchical knowledge-enhanced framework does improve the pre-training performance with the merits of interpretation, which is essential for AI-assisted applications. With this,  clinicians can better understand AI predictions and make collaboration in the computer-aided clinical diagnosis.

\subsection{Generalization} \label{gen}

\subsubsection{Generalization on the open-class setting} \label{gen to brats}
We further verify UniBrain under the open-class setting with BraTS2019. Specifically, we compared the performance of UniBrain on diagnosing HGG and LGG\footnote{Please refer to the dataset description in the previous paragraph for details.} with that of the SOTA method DSM~\cite{chatterjee2022classification} in Table~\ref{brats}. We use the same training and test set reported in DSM paper and thus derive the DSM baseline from the reported results (only F1 score is reported) in the original paper. It is important to note that LGG and HGG on BraTS2019 have not appeared in the categories of our pre-training data, and T1CE modality of BraTS2019 is not used during our pre-training.

In addition, as there are no reports on BraTS2019, we only finetune UniBrain with $\ell_{\text{bce}}$. 

According to Table~\ref{brats}, UniBrain achieved a superior performance compared to KAD~\cite{zhang2023knowledge} under the zero-shot setting. Besides, UniBrain finetuned on BraTS2019 also surpassed DSM~\cite{chatterjee2022classification}, showcasing its ability as a foundation model for open-class diagnosis.

\begin{table*}
\centering
\caption{Comparison of UniBrain with SOTA medical pre-training models on two external in-house datasets. ZS means zero-shot result. The best results are in bold.}
\renewcommand{\arraystretch}{1.0}
\setlength{\tabcolsep}{2.5pt}
\begin{tabular}{ccccccccc}
\hline
\multirow{2}{*}{Method} &  \multicolumn{4}{c}{AHNU} & \multicolumn{4}{c}{WHH}\\
\cmidrule(lr){2-5}\cmidrule(lr){6-9}
  & aAUC & aACC & aF1 & mAP & aAUC & aACC & aF1 & mAP \\
\hline
MedKLIP-ZS~\cite{wu2023medklip} & 62.00 & 54.67 & 39.89 & 33.47 & 59.04 & 48.62 & 37.99 & 30.68 \\
TransMed-ZS~\cite{dai2021transmed} & 67.96 & 62.72 & 41.77 & 36.98 & 65.37 & 61.72 & 41.29 & 34.78 \\
KAD-ZS~\cite{zhang2023knowledge} & 77.29 & 73.25 & 50.52 & 46.38 & 78.44 & 75.34 & 52.48 & 47.58 \\
\hline
\textbf{UniBrain-ZS} & \textbf{86.23} & \textbf{87.03} & \textbf{61.82} & \textbf{62.13} & \textbf{86.50} & \textbf{85.87} & \textbf{62.87} & \textbf{60.09} \\
\hline
\end{tabular}
\label{openset}
\end{table*}

\begin{table*}
\centering
\caption{Comparison between Finetuned UniBrain (termed as UniBrain-FT) and previous SOTA method DSM~\cite{chatterjee2022classification} on BraTS2019 and the zero-shot comparison between UniBrain (marked as UniBrain-ZS) and KAD (marked as KAD-ZS). For DSM, we use the paper's result. The best results are in bold.}
\renewcommand{\arraystretch}{1.1}
\setlength{\tabcolsep}{7pt}
\begin{tabular}{lcccc}
\hline
\multirow{2}{*}{Method} & \multicolumn{4}{c}{BraTS2019} \\
\cmidrule(lr){2-5}
  & aAUC & aACC & aF1 & mAP \\
\hline
DSM~\cite{chatterjee2022classification} & - & - & 90.36 & -  \\
KAD-ZS~\cite{zhang2023knowledge} & 50.52 & 51.48 & 62.54 & 50.27 \\
\hline
\textbf{UniBrain-ZS} & 62.77 & 59.11 & 64.88 & 58.09 \\
\textbf{UniBrain-FT} & \textbf{94.29} & \textbf{93.75} & \textbf{91.97} & \textbf{94.03} \\
\hline
\end{tabular}
\label{brats}
\end{table*}

\subsubsection{Generalization on the domain shift setting}

To study the performance of UniBrain under the domain shift setting, we verified the pre-training models on AHNU and WHH which have same categories as SSPH but with the covariate shift. We test the zero-shot results to assess the generalization ability of UniBrain compared with two VLP models and TransMed. Table~\ref{openset} suggests the remarkable potential of UniBrain across different hospitals, which can be a promising prospect.

\section{Ablation Study} \label{ablation study}
In this section, we conduct a thorough ablation study for UniBrain to understand the performance impact of the individual components, namely, ARD, visual-language encoders and hierarchical knowledge-enhancement. The ablation results of different components are summarized in Table~\ref{ablation}, which we will explain in details for each component in the following.

\subsection{Effectiveness of Automatic Report Decomposition}
\subsubsection{Effectiveness of hierarchical report extraction}
To show the effectiveness of the proposed ARD, we compare it with ChatGPT~\cite{brown2020language}. Specially, we provide the predefined decomposition examples as demonstrations to ask ChatGPT to perform the report decomposition\footnote{An complete prompt follows this format: \scriptsize{\textsf{Report decomposition introduction}} $\textbf{\#}$ \scriptsize{\textsf{Decomposition demonstrations}} $\textbf{\#}$ \scriptsize{\textsf{Reports to be decomposed}}.}, and provide the results in the rows \textit{w.r.t.} ``w/ chatGPT report" of Table~\ref{ablation}. As can be seen, compared to ChatGPT~\cite{brown2020language}, ARD shows the superior performance with increases in aAUC, aACC, aF1, and mAP by 0.93\%, 0.15\%, 0.75\%, and 1.36\% respectively. It suggests that the proper human experience and labor in the extraction make the report decomposition more informative and efficient.
\subsubsection{Effectiveness of disease label extraction}
As aforementioned in previous section, we directly generate the label for each MRI sample by checking whether the report impression contains the corresponding pathology. We verify our generation process with an evaluation set of 250 samples. In this set, experienced radiologists provided the gold standard disease labels for each imaging-report pair. The performance of ARD in disease label extraction was evaluated by comparing with the gold standard labels on the evaluation set, under on the tasks of mention and negation detection measured by the F1 score. Specially, mention detection should give a positive label if the gold standard shows the patient gets the certain disease, while negation detection should give a positive label if the gold standard shows the patient does not have the certain disease. The results in Table~\ref{label_acc} reveal that ARD reaches a competitive precision in extracting disease labels from report impression.

\subsection{Performance under Different Encoders}
In UniBrain, we adopted ResNet3D 34 as the image encoder. Here, we verify the performance of ResNet3D 50 to see how network depth affects the results. Regarding the text encoder, we substitute MedKEBERT~\cite{zhang2023knowledge} in UniBrain with clinicalBERT which is pre-trained on clinical notes~\cite{huang2019clinicalbert}, to validate its efficiency. As shown in the rows \textit{w.r.t.} encoders of Table~\ref{ablation} (w/ ResNet3D 50, w/ clinicalBERT), we can find that the depth 50 of image encoder does not show the significant different with the depth 34 in UniBrain. However, changing text encoder to clinicalBERT~\cite{huang2019clinicalbert} leads to the degraded performance, since UMLS is a more structured knowledge corpora than clinical notes. This suggests pre-training the text encoder with the well-organized knowledge base is critical to enhance the VLP.

\begin{table*}
\centering
\caption{The label extraction performance of ARD on the test set for mention and negation detection. The Macro-average and Micro-average rows are computed over all 13 classes.}
\renewcommand{\arraystretch}{1.1}
\setlength{\tabcolsep}{3pt}
\begin{tabular}{ccc}
\hline
Category &  \makecell{Mention F1} & \makecell{Negation F1}\\
\hline
Lacunar cerebral infarction & 100.00 & 100.00  \\
Brain atrophy & 100.00 & 100.00 \\
White matter lesions & 100.00 & 100.00 \\
Chronic cerebral infarction & 100.00 & 100.00 \\
Normal & 100.00 & 100.00 \\
Metastasis & 100.00 & 100.00 \\
Hydrocephalus & 100.00 & 100.00 \\
Brain hemorrhage & 81.25 & 98.74 \\
Acute cerebral infarction & 72.72 & 98.77 \\
Epidura and subdural hemorrhage & 96.43 & 99.56 \\
Meningioma & 95.65 & 99.57 \\
Hemangioma & 97.77 & 99.78 \\
Glioma & 92.31 & 99.79 \\
\hline
Micro-average & 97.32 & 99.70 \\
Macro-average & 95.09 & 99.71 \\
\hline
\end{tabular}
\label{label_acc}
\end{table*}

\begin{table*}[!t]
\footnotesize
\centering
\renewcommand{\arraystretch}{1.2}
\caption{Ablation study of UniBrain. The best results are in bold.}
\begin{tabular}{lp{0.5cm}<{\centering}p{0.5cm}<{\centering}p{0.5cm}<{\centering}p{0.5cm}<{\centering}p{0.5cm}}
\hline
Methods & aAUC & aACC & aF1 & mAP \\ 
\hline
Ablation of ARD & & & & \\
w/ chatGPT report & 89.78 & 92.47 & 61.52 & 61.91 \\
\hline
Ablation of encoders & & & & \\
w/ clinicalBERT~\cite{huang2019clinicalbert} & 89.63 & 92.11 & 59.59 & 59.81\\
w/ ResNet3D 50 & 90.62 & 92.77 & 62.00 & 62.86 \\
\hline
On hierarchical knowledge-enhancement & & & & \\
w/o all imaging-report alignments & 89.27 & 92.27 & 61.46 & 61.50\\
w/o modality-wise alignment & 89.87 & 92.07 & 61.40 & 60.68 \\
w/o disease description query & 84.62 & 83.64 & 51.73 & 48.29 \\
w/o CVP & 89.85 & 92.27 & 60.61 & 60.66\\
\hline
\textbf{UniBrain} & \textbf{90.71} & \textbf{92.62} & \textbf{62.27} & \textbf{63.27} \\
\hline
\end{tabular}
\footnotetext{AUC, MCC, F1 and ACC scores are reported, 
and the metrics all refer to the macro average on all the diseases. Numbers within parentheses indicate $95\%$ CI. The best results are bold.}
\label{ablation}
\end{table*}

\subsection{On Hierarchical Knowledge-enhancement} \label{effecthk}
\subsubsection{Effectiveness of hierarchical imaging-report alignment}
As shown in Table~\ref{ablation}, we can observe a significant drop in performance when all imaging-report alignments are absent, \textit{i.e.} eliminating $\ell_\text{g}$ and all $\ell_k$ in Eq.~\eqref{eq:overall}. It suggests that the process of knowledge-enhancement helps learn better cross-vision-language representations, resulting in the improvement. Furthermore, when only keeping the global imaging-report alignment (\textit{i.e.}, w/o modality-wise alignment), the results are better than removing all imaging-report alignments but worse than the complete UniBrain. This indicates the effectiveness of the hierarchical imaging-report alignment during pre-training.

\subsubsection{Effectiveness of the coupled vision-language perception}
Finally, we verified the effect of the CVP module in UniBrain. According to Table~\ref{ablation}, removing CVP degrades the performance (w/o CVP) significantly in aF1 and mAP, which shows the importance of CVP in transforming the global image features into disease-specific classification features. Besides, we explore the impact of the query form for CVP by conducting an additional experiment with using the disease name instead of the disease description as $Q$. From the results of Table~\ref{ablation} \textit{w.r.t.,} w/o disease description query, we can find that the model using the disease description as the query significantly outperforms the model using the disease name. We attribute this to the fact that the disease description provides more comprehensive information, allowing the text feature to pay more robust attention on the image features.

\section{Discussion}
\label{discussion}

\subsection{Clinical Impact Analysis}
UniBrain is primarily dedicated to more general brain disease diagnosis, and our experiments demonstrate that UniBrain outperforms the current state-of-the-art methods in diagnosis accuracy and interpretation. We consider UniBrain as a promising avenue for computer-aided diagnosis in brain MRI, potentially enhancing the productivity and alleviating the workload of radiologists. Additionally, the adaptable input query length of CVP enables UniBrain to seamlessly extend to a range of downstream tasks. Experiments in Section~\ref{gen} shows the competitive performance even in challenging scenarios involving open-class and domain shift settings, thus broadening the scope of UniBrain in clinical applications.

\begin{figure}[t!]
\centerline{\includegraphics[width=\linewidth,height=.25\linewidth]{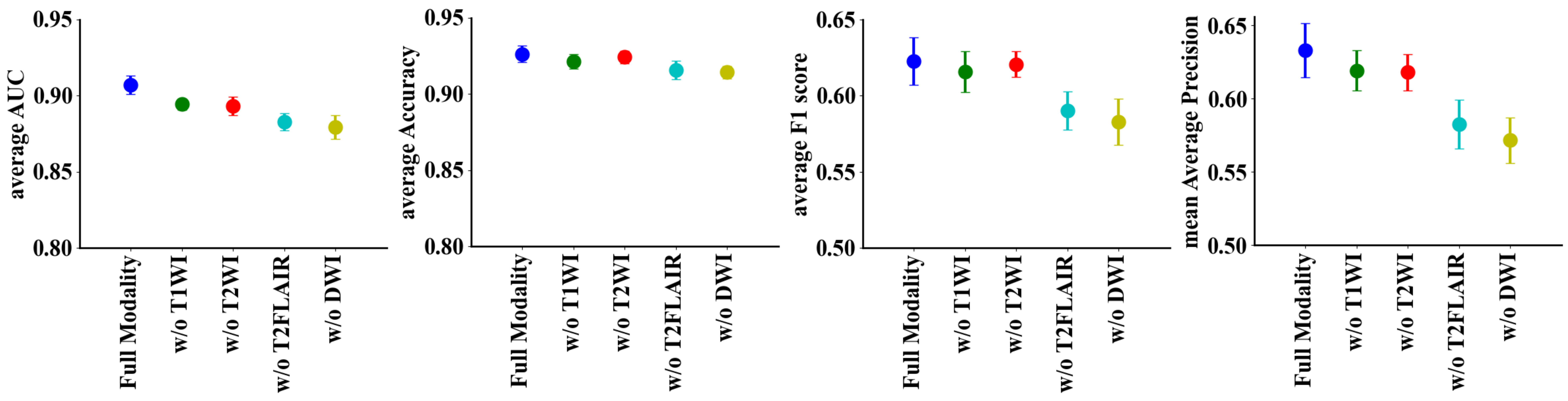}}
\caption{Comparison of full-modality with modality absence situations in terms of aAUC, aACC, aF1 and mAP.}
\label{fig:absence}
\end{figure}

\subsection{Limitations and Future Work}
Despite the effectiveness, there remains a few limitations to UniBrain: 1) the report primarily comprises abnormal observations, while the image contains both the normal and abnormal patches. Consequently, the current alignment of imaging and report may lead to the overfitting between normal image patches with the abnormal observations in the report. More explorations can take this point into consideration in the future to improve the alignment mechanism; 2) UniBrain can be greatly affected by the modality absence as shown in Fig.~\ref{fig:absence}, which is a common problem in the real-world scenarios. As it is hard to guarantee the full modalities in some cases, it will be important to consider the robustness of the pre-training models to handle this problem. 3) UniBrain lacks the capability of the pixel-level lesion segmentation. As a prospective direction, the objective is to develop an integrated brain MRI diagnosis system that entails both classification and pixel-level lesion segmentation capabilities for multiple types of diseases.

\section{Conclusion}
\label{conclusion}
In this work, we propose a hierarchical knowledge-enhanced pre-training framework named UniBrain for the general diagnosis of brain disorders. First, we design an automatic report decomposition method to efficiently extract the important pieces of report for pre-training. Second, we perform a hierarchical imaging-report alignment to achieve a fine-grained knowledge enhancement. Lastly, we input the global image features and disease query set into the coupled vision-language perception module to generate the final diagnosis and grounding. Experiments show UniBrain not only outperforms SOTA methods on both in-house and public datasets  under open-class and domain shift settings but also yields comparable performance on certain categories compared to human experts.

\bibliographystyle{plainnat}
\bibliography{references}
\clearpage

\end{document}